\documentclass{article}

\usepackage{times}

\usepackage[square,numbers]{natbib}
\usepackage{multicol}
\usepackage[bookmarks=true]{hyperref}
\usepackage{graphicx}
\usepackage{capt-of}
\usepackage{caption}

\usepackage{amsmath} 
\usepackage{amssymb}  
\usepackage{balance}
\usepackage{times}
\usepackage{dsfont}
\usepackage{siunitx}
\usepackage{cleveref}
\usepackage{algorithm}
\usepackage{algpseudocode}
\usepackage{algorithmicx}
\usepackage{wrapfig}
\usepackage{xspace}
\usepackage{xurl}
\usepackage[usenames,dvipsnames,table,xcdraw]{xcolor}
\definecolor{lightgray}{gray}{0.9}
\usepackage{booktabs}  
\usepackage{float} 
\usepackage[flushleft]{threeparttable}
\usepackage{amsmath,amssymb,amsfonts}
\usepackage{arydshln} 

\usepackage{setspace}

\usepackage{stfloats}

\definecolor{mydarkblue}{rgb}{0,0.08,0.45}
\definecolor{mydarkgreen}{RGB}{0, 139, 69}
\definecolor{mygreen2}{RGB}{0 205 0}
\definecolor{mybrown}{RGB}{139 69 19}

\definecolor{boxblue}{RGB}{79,173,234}
\definecolor{boxgreen}{RGB}{159,206,99}

\definecolor{tablepeach}{RGB}{255, 240, 235}
\definecolor{tablepurple}{RGB}{248,235,252}
\definecolor{tableblue}{RGB}{235,241,255}

\hypersetup{
	colorlinks=true,
	linkcolor=blue,
	urlcolor=mybrown,
	citecolor=mygreen2,
}

\newcommand{\paragraphc}[1]{\vspace{0.2em}\noindent\textbf{#1}}

\usepackage[final]{corl_2025} 

\title{Sim-to-Real Reinforcement Learning for \\ Vision-Based Dexterous Manipulation on Humanoids}

\author{
Toru Lin${}^{1,2}$ \quad
Kartik Sachdev${}^{2}$ \quad
Linxi ``Jim'' Fan${}^{2}$ \quad
Jitendra Malik${}^{1}$ \quad
Yuke Zhu${}^{2,3}$ \quad \\ \\
UC Berkeley${}^{1}$ \quad NVIDIA${}^{2}$ \quad UT Austin${}^{3}$\\
\\
\url{https://toruowo.github.io/recipe} \\
}

\begin{document}
\maketitle


\vspace{-2.5em}
\begin{figure}[!ht]
    \includegraphics[width=\linewidth]{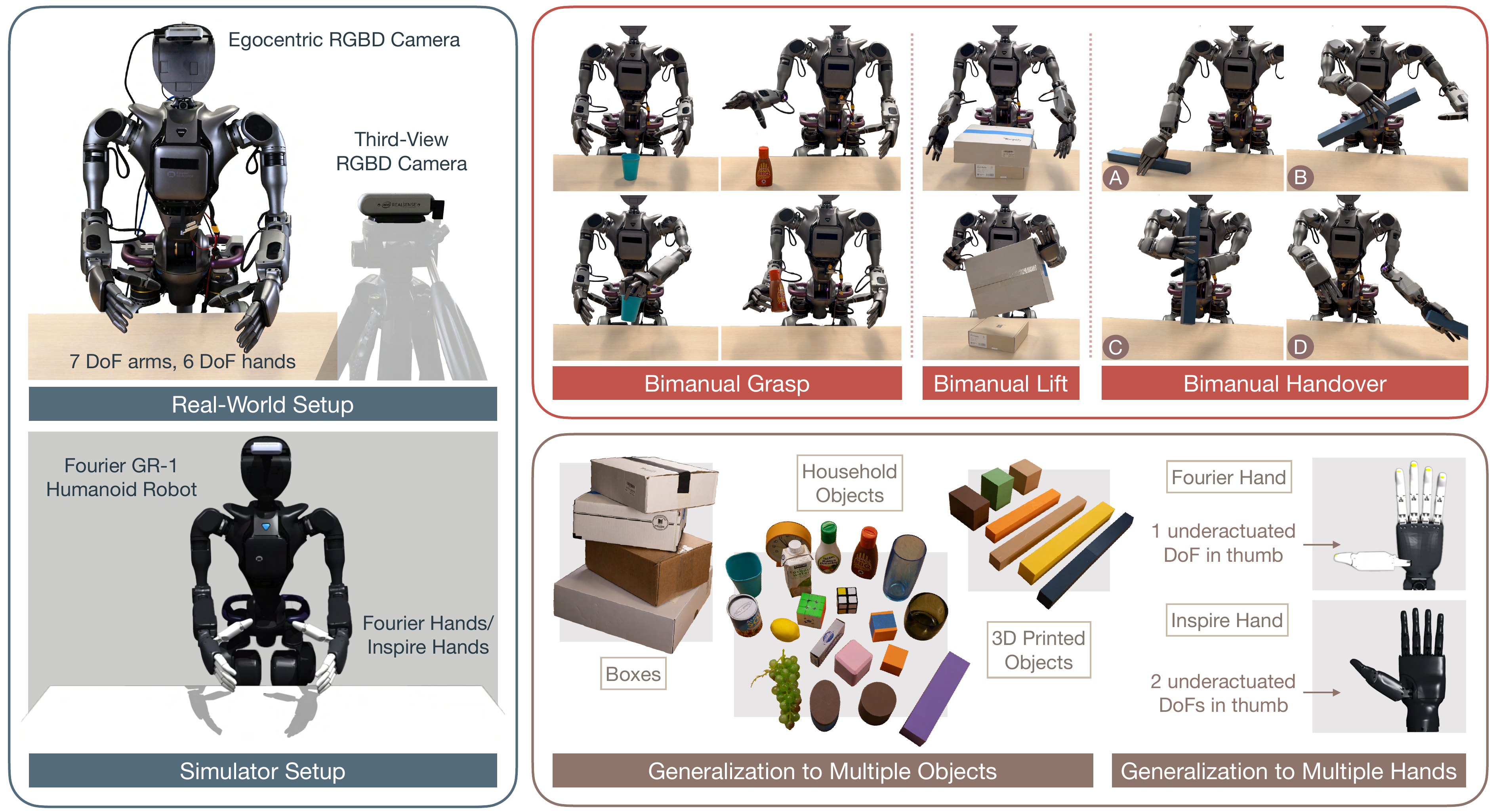}
    \caption{\small{\textbf{Overview.} We train a humanoid robot with two multi-fingered hands to perform a range of contact-rich dexterous manipulation tasks on diverse objects. Observations are obtained from a third-view camera, an egocentric camera, and robot proprioception. Our reinforcement learning policies generalize zero-shot to unseen real-world objects with varying physical properties (e.g. shape, size, color, material, mass) and remain robust against force disturbances. We also validate the adaptability of our approach on two hardware variations.
    }}
    \vspace{-0.1em}
    \label{fig:teaser}
\end{figure}


\vspace{-1em}
\begin{abstract}
    Learning generalizable robot manipulation policies, especially for complex multi-fingered humanoids, remains a significant challenge. Existing approaches primarily rely on extensive data collection and imitation learning, which are expensive, labor-intensive, and difficult to scale. Sim-to-real reinforcement learning (RL) offers a promising alternative, but has mostly succeeded in simpler state-based or single-hand setups. How to effectively extend this to vision-based, contact-rich bimanual manipulation tasks remains an open question.
    In this paper, we introduce a practical sim-to-real RL recipe that trains a humanoid robot to perform three challenging dexterous manipulation tasks: grasp-and-reach, box lift and bimanual handover.
    Our method features an automated real-to-sim tuning module, a generalized reward formulation based on contact and object goals, a divide-and-conquer policy distillation framework, and a hybrid object representation strategy with modality-specific augmentation.
    We demonstrate high success rates on unseen objects and robust, adaptive policy behaviors -- highlighting that vision-based dexterous manipulation via sim-to-real RL is not only viable, but also scalable and broadly applicable to real-world humanoid manipulation tasks.
\end{abstract}



\section{Introduction}
\label{sec:intro}
\vspace{-1em}

Learning generalizable manipulation policies -- especially for complex humanoid robots equipped with multi-fingered hands -- remains a formidable challenge in robotics. Existing approaches often rely on extensive real-world data collection and imitation learning~\cite{bjorck2025gr00t,cheng2024open,lin2024learning}, which are costly, labor-intensive, and difficult to scale. Sim-to-real reinforcement learning (RL) offers a promising alternative and has achieved impressive results in navigation~\cite{haarnoja2024learning,rudin2022advanced}, locomotion~\cite{hwangbo2019learning,lee2020learning}, and autonomous drone racing~\cite{kaufmann2023champion}. However, its application to dexterous manipulation remains largely limited to single-hand~\cite{akkaya2019solving,handa2023dextreme,chen2022system,qi2023general,singh2024dextrah} or state-based setups~\cite{lin2024twisting,huang2023dynamic,qi2022hand,lum2024dextrah}, leaving open the question of how to scale RL to vision-based, contact-rich bimanual tasks on humanoid embodiments.

In this work, we present a practical vision-based sim-to-real RL recipe that enables a multi-fingered humanoid robot to learn highly generalizable, robust, and dexterous manipulation skills. We identify and address several key challenges that have not been thoroughly explored in prior works:

\vspace{-1mm}
\textbf{(A) Sim-to-real for low-cost manipulation systems.} Existing approaches rely on industry-grade robotic arms with high-precision motors. However, many humanoid platforms employ much more lightweight, noisier motors. This makes contact-rich dexterous grasping and bimanual coordination significantly harder, especially with sim-to-real method. We introduce a simple, automated real-to-sim system identification method to overcome this with less than four minutes of real-world data.

\vspace{-2mm}
\textbf{(B) Reward design for complex coordination.} Bimanual manipulation tasks such as handover and lifting require complex coordination between arms and hands: one side must act in a way that complements the other, with precision in both motion and timing. Designing a reward function that captures this type of contact-rich collaboration is nontrivial. We propose a novel keypoint-based reward formulation to facilitate such coordination effectively.

\vspace{-2mm}
\textbf{(C) Exploration.} The long-horizon, high-dimensional nature of bimanual coordination introduces a hard exploration problem, even when reward functions are well-shaped. We propose to use a task-aware initialization strategy to accelerate single task RL, and decompose the overall multi-task~(e.g. object) policy learning into separate single-task RL followed by generalist policy distillation.  

\vspace{-2mm}
\textbf{(D) Object perception} The combination of object diversity and sim-to-real domain shift makes vision-based manipulation particularly difficult. We propose a hybrid object representation that combines compact low-dimensional features with expressive high-dimensional features, augmented via modality-specific randomization. We find this simple strategy surprisingly effective -- improving sim-to-real success rates on novel objects significantly by 80$\sim$100\%.

We demonstrate the effectiveness of our approach on three challenging vision-based manipulation tasks: dexterous grasp-and-reach, bimanual lifting, bimanual handover. Our zero-shot sim-to-real policies exhibit robust, adaptive, and generalizable behavior on unseen real-world objects with diverse physical properties, achieving 90\% success rate on seen objects and 60$\sim$80\% success rate on novel objects. Additionally, we confirm our method's adaptability to hardware variation across two distinct multi-fingered robot hands. Together, these results establish a practical and scalable recipe for high-performance vision-based dexterous manipulation via sim-to-real RL.

\vspace{-0.5em}
\section{Background}
\vspace{-1em}

\subsection{Deep Reinforcement Learning Applications to Robotics}
\vspace{-1em}

The successes of deep RL across domains like gaming, language modeling, and control~\cite{hwangbo2019learning,kaufmann2023champion,openai2024,deepseekai2025deepseekr1incentivizingreasoningcapability,silver2017mastering,vinyals2019grandmaster} have generated widespread excitement. However, the paradigm is known to be brittle, with sensitivity to hyperparameters~\cite{henderson2018deep} and reproducibility issues~\cite{islam2017reproducibility} due to high algorithmic variance.

Among open problems in RL, exploration remains fundamental. Unlike supervised learning, RL agents must collect their own data — and the strategy for doing so directly affects performance. Real-world robotics compounds this difficulty with high-dimensional inputs, sparse rewards, and complex dynamics. Numerous methods have aimed to scale exploration by incentivizing novelty~\cite{bellemare2016unifying,burda2018exploration,lin2024mimex,ostrovski2017count,pathak2017curiosity,stadie2015incentivizing,tang2017exploration}, but they do not fundamentally resolve the exploration bottleneck.

Robotics also exposes challenges overlooked in standard RL benchmarks~\cite{bellemare2013arcade,tassa2018deepmind}, including: (1) the absence of fully modeled environments, and (2) the lack of clearly defined reward functions. Past works have introduced practical techniques to mitigate these issues, such as learning from motion capture or teleoperated demonstrations~\cite{chen2024object,rajeswaran2017learning,yin2025dexteritygen,zhu2018reinforcement}, real-to-sim modeling techniques~\cite{akkaya2019solving,haarnoja2024learning,handa2023dextreme,lin2024twisting,torne2024rialto}, and more principled reward design~\cite{memmel2024asid,zhang2024wococo}. While often tailored to specific tasks or hardware, these approaches lay groundwork that our method builds upon and generalizes.

\subsection{Vision-Based Dexterous Manipulation on Humanoids}
\vspace{-1em}

\paragraphc{Imitation learning and classical approaches.}
Recent advances in teleoperation~\cite{cheng2024open,lin2024learning,wu2023gello,zhao2023learning} and learning from demonstrations~\cite{chi2023diffusion,li2024planning} have enabled significant progress in vision-based dexterous manipulation~\cite{cheng2024open,li2024planning,lin2024learning,zhao2024aloha}. However, teleoperation remains costly to scale, and achieving high success rates with real-world demonstration data alone~\cite{levine2018learning,lin2024data,zhao2024aloha} requires large datasets, making purely supervised methods expensive for reaching human-level performance on complex tasks.

\paragraphc{Reinforcement learning approaches.} 
RL-based manipulation works have shown strong results in settings such as in-hand reorientation~\cite{akkaya2019solving,handa2023dextreme,qi2023general,wang2024lessons}, grasping~\cite{lum2024dextrah,singh2024dextrah}, twisting~\cite{lin2024twisting}, and dynamic handover~\cite{huang2023dynamic}, but typically focus on single-hand setups~\cite{akkaya2019solving,chen2023sequential,handa2023dextreme,lum2024dextrah,qi2023general,singh2024dextrah,wang2024lessons} or use intermediate object representations rather than raw pixels~\cite{chen2024object,huang2023dynamic,lin2024twisting}. The closest to our work is Chen et al.~\cite{chen2024object}, but their method relies on human hand motion capture, while our work learns full hands-arms joint control from scratch. Our work is also the first to demonstrate robust sim-to-real transfer of bimanual policies on a novel humanoid platform with multi-fingered hands.


\begin{figure*}[t]
\begin{center}
\includegraphics[width=\textwidth]{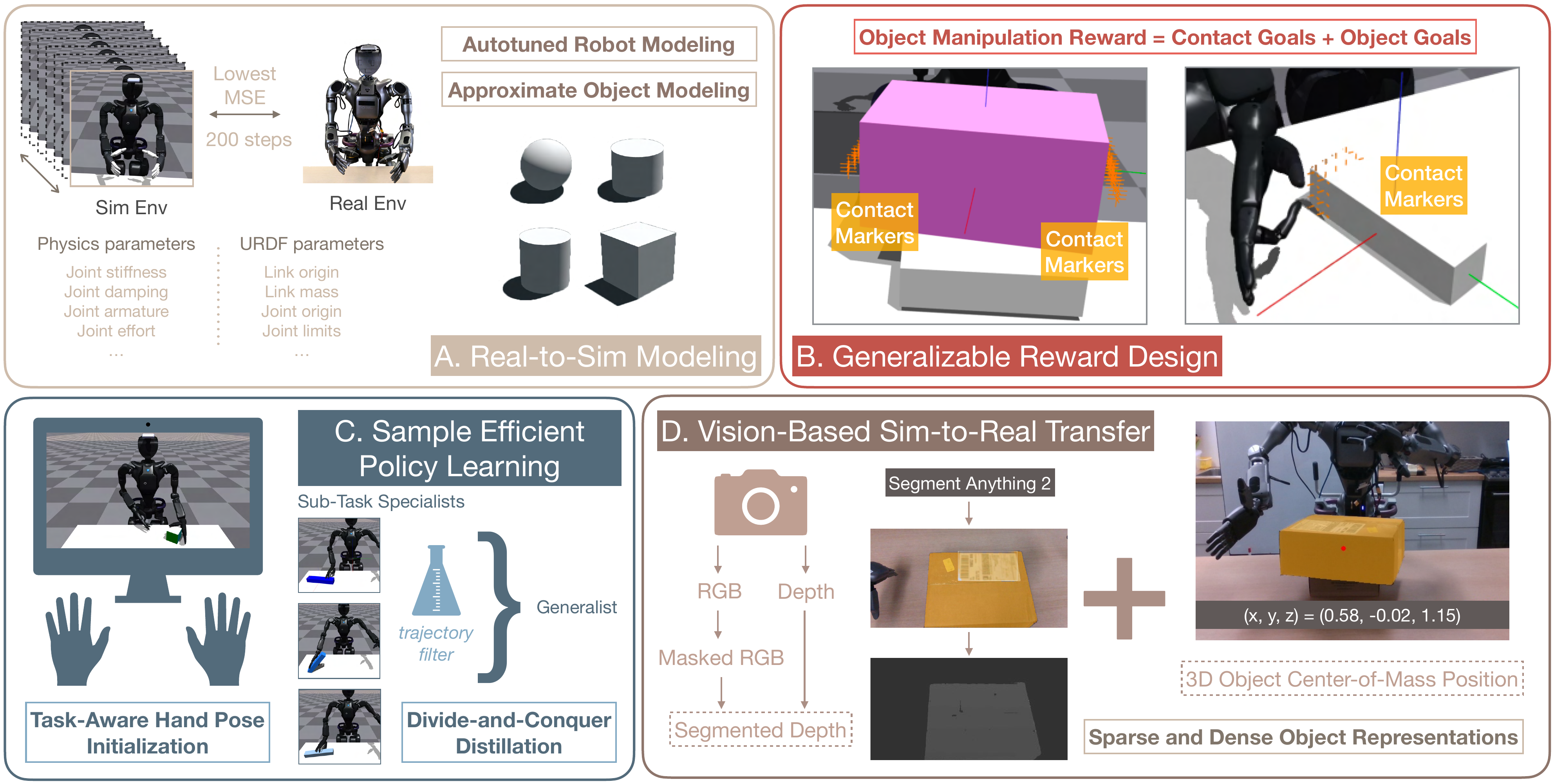}
\end{center}
\vspace{-1em}
\caption{\small{\textbf{A sim-to-real RL recipe for vision-based dexterous manipulation.} We close the environment modeling gap between simulation and reality through an automated real-to-sim tuning module, design generalizable task rewards by disentangling each manipulation task into contact states and object states, improve sample efficiency of policy training by using task-aware hand poses and divide-and-conquer distillation, and transfer vision-based policies to the real world with a mixture of sparse and dense object representations.}}
\label{fig:overview}
\vspace{-1em}
\end{figure*}


\section{Our Recipe}
\vspace{-1em}

Section~\ref{sec:intro} outlined four key challenges in sim-to-real RL for dexterous manipulation. Here, we provide the detailed approaches we develop for each. An overview is shown in Figure~\ref{fig:overview}.

\subsection{Real-to-Sim Modeling}
\label{sec:realsim}
\vspace{-1em}

Simulators offer unlimited trial-and-error chances to perform the exploration necessary for RL. However, whether policies learned in simulation can be reliably transferred to the real world hinges on accurate modeling of both robots and environments. In dexterous manipulation, this challenge is compounded by the necessity to model objects, which have diverse and often unmeasurable physical properties. Even with known parameters, matching real-world and simulated dynamics is nontrivial: the same values for physical constants in simulation and the real world do not necessarily correspond to identical kinematic and dynamic relationships due to discrepancies in physics engines.

\begin{wrapfigure}[24]{r}{0.5\textwidth}  
\vspace{-1em}
\begin{minipage}{0.5\textwidth}
\begin{algorithm}[H]
\small
\setstretch{0.95}
\caption{\small{Real-to-Sim Autotune Module}}
\begin{algorithmic}[1]
\Require
    \State $E$ : Set of environment parameters to tune
    \State $N$ : Number of calibration action sequences
    \State $R$ : Real robot hardware environment
    \State $M$ : Initial robot model file
    
\Procedure{Autotune}{$E, N, R, M$}
    \State $P \gets \text{InitializeParameterSpace}(E, M)$ 
    \State $S \gets \{\}$ \Comment{Set of simulated environments}
    
    \For{$i \gets 1$ to $K$}  \Comment{$K$ is population size}
        \State $p_i \gets \text{RandomSample}(P)$
        \State $S_i \gets \text{CreateSimEnvironment}(p_i)$
        \State $S \gets S \cup \{S_i\}$
    \EndFor

    \State $J \gets \text{GenerateJointTargets}(N)$ 
    \State $R_{track} \gets \text{GetTrackingErrors}(R, J)$ \Comment{Real tracking}

    \State $best\_params \gets \text{null}$
    \State $min\_error \gets \infty$
    
    \For{$S_i \in S$}
        \State $S_{track} \gets \text{GetTrackingErrors}(S_i, J)$
        \State $error \gets \text{ComputeMSE}(S_{track}, R_{track})$
        \If{$error < min\_error$}
            \State $min\_error \gets error$
            \State $best\_params \gets \text{GetParameters}(S_i)$
        \EndIf
    \EndFor
    
    \Return $best\_params$
\EndProcedure
\end{algorithmic}
\label{alg:autotune}
\end{algorithm}
\end{minipage}
\vspace{-2em}
\end{wrapfigure}

\paragraphc{Autotuned robot modeling.} Manufacturer-supplied robot models offer a baseline, but often require significant tuning~\cite{akkaya2019solving,radosavovic2024real} to be ready for sim-to-real. This tuning is a laborious process as there is no ``ground truth'' pairing between the real world and the simulated world. We propose an \textit{autotune} module for fast, automated calibration of simulation parameters to match real robot behavior. 
As shown in Figure~\ref{fig:overview}A and Algorithm~\ref{alg:autotune}, our method jointly optimizes simulator physics (e.g. friction, damping) and URDF constants (e.g. link inertia values, joint limits) using only a single set of calibration trajectories on real hardware. It samples parameter sets, runs joint-targeted motions in parallel simulations, and selects the set minimizing tracking error against the real robot -- automatically searching the parameter space to identify optimal values for both simulator physics and robot model constants in under four minutes (or 2000 simulated steps in \SI{10}{\Hz}). This removes the need for iterative manual tuning and generalizes to any simulator-exposed parameter affecting kinematic behaviors.

\paragraphc{Approximate object modeling.} 
Following prior work~\cite{lin2024twisting,qi2023hand}, we model objects using simple geometric primitives (e.g. cylinders) with randomized physical parameters. Despite their simplicity, these approximations are sufficient to learn dexterous manipulation policies that transfer reliably to the real world. Our recipe adopts this strategy and finds it both effective and generalizable.

\vspace{-1em}
\subsection{Generalizable Reward Design}
\label{sec:reward}
\vspace{-1em}

In standard RL~\cite{sutton1998introduction}, the reward function plays a central role in shaping agent behavior. However, much of RL research has treated rewards as fixed, focusing instead on algorithmic improvements~\cite{eschmann2021}. In robotics — and especially in dexterous manipulation — designing effective, generalizable rewards becomes a key challenge due to complex contact dynamics and object variability~\cite{dewey2014reinforcement}.

\paragraphc{Manipulation as contact and object goals.} 
We observe that many human manipulation tasks~\cite{grauman2024ego} can be decomposed into a sequence of hand-object contact transitions and object state changes. Inspired by this, we propose a structured reward design scheme for long-horizon, contact-rich tasks. For instance, a bimanual handover can be segmented into: (1) one hand contacting the object, (2) lifting the object near the second hand, (3) the second hand contacting the object, and (4) transferring the object to the target location. We therefore define rewards based on two key components: ``contact goals'' encourages the fingertips to reach task-relevant contact points on object, and ``object goals'' penalizes current object state deviation from the target object state (e.g. \textit{xyz} position). To facilitate contact goal specification, we introduce a keypoint-based technique: simulated objects are augmented with ``contact stickers'' — surface markers representing desirable contact locations. The contact goal, in terms of reward, can then be specified as $r_{\mathrm{contact}} = \sum_{i}\left[\frac{1}{1+\alpha d(\mathbf{X}^L, \mathbf{F}^L_i)} + \frac{1}{1+\beta d(\mathbf{X}^R, \mathbf{F}^R_i)}\right]$, where $\mathbf{X}^L\in\mathbb{R}^{n\times 3}$ and $\mathbf{X}^R\in\mathbb{R}^{m\times 3}$ are the positions of contact markers specified for left and right hands, $\mathbf{F}^L\in\mathbb{R}^{4\times 3}$ and $\mathbf{F}^R\in\mathbb{R}^{4\times 3}$ are the position of left and right fingertips, $\alpha$ and $\beta$ are scaling hyperparameters, and $d$ is a distance function defined as $d(\mathbf{A}, \mathbf{x}) = \min_i \Vert\mathbf{A}_i - \mathbf{x}\Vert_2$.
These contact markers can be arbitrarily specified -- for example, procedurally generated based on object geometry -- offering a flexible way to incorporate contact preferences or human priors.
A visualization of contact markers is shown in Figure~\ref{fig:overview}B, and their empirical effectiveness is analyzed in Section~\ref{sec:exp}.

\subsection{Sample Efficient Policy Learning}
\label{sec:policy}
\vspace{-1em}

Even with a well-shaped reward, learning dexterous policies on high-dimensional bimanual multi-fingered systems remains sample-inefficient due to sparse rewards and exploration complexity. We introduce two techniques to improve sample efficiency: (1) task-aware initialization using human-guided hand poses, and (2) a divide-and-conquer strategy with policy distillation.

\paragraphc{Task-aware hand poses for initialization.} We collect task-relevant hand-object configurations from human teleoperation in simulation. This can be done using any compatible system for bimanual multi-fingered hands. The recorded states, including object poses and robot joint positions, are then randomly sampled as initial conditions for each training episode. Unlike prior work that relies on full demonstration trajectories~\cite{bauza2024demostart}, our approach only requires humans to casually ``play around'' with the task goal in mind. This lightweight data collection takes less than 30 seconds per task since no expert demonstration is needed, yet proves highly effective in improving early-stage exploration.

\paragraphc{Divide-and-conquer distillation.} Standard RL exploration techniques~\cite{burda2018exploration,lin2024mimex,pathak2017curiosity,taiga2019benchmarking} aim to visit the state space more efficiently but do not fundamentally alter the difficulty of sparse-reward problems: the probability of receiving learning signals from visiting the ``right'' states remains the same. We instead overcome the exploration problem by breaking down the explorable state space itself, e.g. decomposing a multi-object manipulation task into multiple single-object manipulation tasks. Once specialized policies are trained for each sub-task, high-quality rollouts can be filtered and distilled into a generalist policy using shared observation and action spaces. This effectively brings pure RL closer to learning from demonstrations, where the sub-task policies act as ``teleoperators'' in the simulation environment, and the centralized generalist policy learns from curated data.

\subsection{Vision-Based Sim-to-Real Transfer}
\label{sec:vis}
\vspace{-1em}

Vision-based sim-to-real transfer is particularly challenging due to domain gaps in both dynamics and perception. We employ two strategies to address these challenges: hybrid object representations and extensive domain randomization.

\paragraphc{Hybrid object representations.} Dexterous manipulation often requires precise perception of object pose and geometry. Prior work spans a spectrum of object representations, from 3D position~\cite{lin2024twisting} and 6D pose~\cite{akkaya2019solving}, to depth~\cite{lum2024dextrah,qi2023general}, point cloud~\cite{liu2024visual}, and RGB images~\cite{handa2023dextreme}. Higher-dimensional representations encode richer information about the object, improving task performance but also widening the sim-to-real gap; and vice versa. To balance the trade-offs, we propose to use a mix of low- and high-dimensional signals: 3D object position (from a fixed third-person view) and depth image (from an egocentric view). We obtain the 3D object position from a reliable object tracking module with relatively controllable noise, and use segment out the object depth to reduce the visual sim-to-real gap. We validate this design in Section~\ref{sec:exp}.

\paragraphc{Domain randomization for perception and dynamics.} To improve robustness, we apply extensive domain randomization during training. This includes variation in object parameters, camera parameters, robot physical properties, and observation noises. Full details are provided in Appendix~\ref{sec:dr}.

\section{Experiments}
\label{sec:exp}
\vspace{-1em}

Our proposed approaches form a general recipe that allows for the practical application of RL to solve dexterous manipulation with humanoids. In this section, we show experimental results of task capabilities and ablation studies of each proposed technique. Videos can be found on our website.

\subsection{Real-World and Simulator Setup}
\vspace{-1em}
We use a Fourier GR1 humanoid robot with two arms and two multi-fingered hands. Each arm has 7 degrees of freedom (DoF). For most experiments, we use the Fourier hands, each of which has 6 actuated DoFs and 5 underactuated DoFs. To show cross-embodiment generalization, we include results on the Inspire hands, each with 6 actuated DoFs and 6 underactuated DoFs. The hardware has substantially different masses, surface frictions, finger and palm morphologies, and thumb actuations. Figure~\ref{fig:teaser} visualizes both hands.
We use the NVIDIA Isaac Gym simulator~\cite{makoviychuk2021isaac}.

\paragraphc{Perception.} As outlined in Section~\ref{sec:vis}, we use a combination of dense and sparse object features for policy learning in both simulation and real-world transfer. In the real world, we set up an egocentric-view RealSense D435 depth camera on the head of the humanoid robot and a third-view RealSense D435 depth camera on a tripod in front of the robot (illustrated in Figure~\ref{fig:teaser}). In simulation, we similarly set up the two cameras by calibrating their poses against the real camera poses. The \textit{dense} object feature is obtained by directly reading depth observations from the egocentric-view camera. The \textit{sparse} feature is obtained by approximating the object's center-of-mass from the third-view camera, using a similar technique as in Lin et al.~\cite{lin2024twisting}. As illustrated in Figure~\ref{fig:overview}, we use the Segment Anything Model 2 (SAM2)~\cite{ravi2024sam} to generate a segmentation mask for the object at each trajectory sequence's initial frame, and leverage the tracking capabilities of SAM2 to track the mask throughout all remaining frames. To approximate object's 3D center-of-mass coordinates, we calculate the center position of object mask in the image plane, then obtain noisy depth readings from a depth camera to recover a corresponding 3D position. The perception pipeline runs at \SI{5}{\Hz} to match the neural network policy's control frequency.

\begin{figure*}[t]
  \centering
  \begin{minipage}[t]{0.36\textwidth}
    \includegraphics[width=\linewidth]{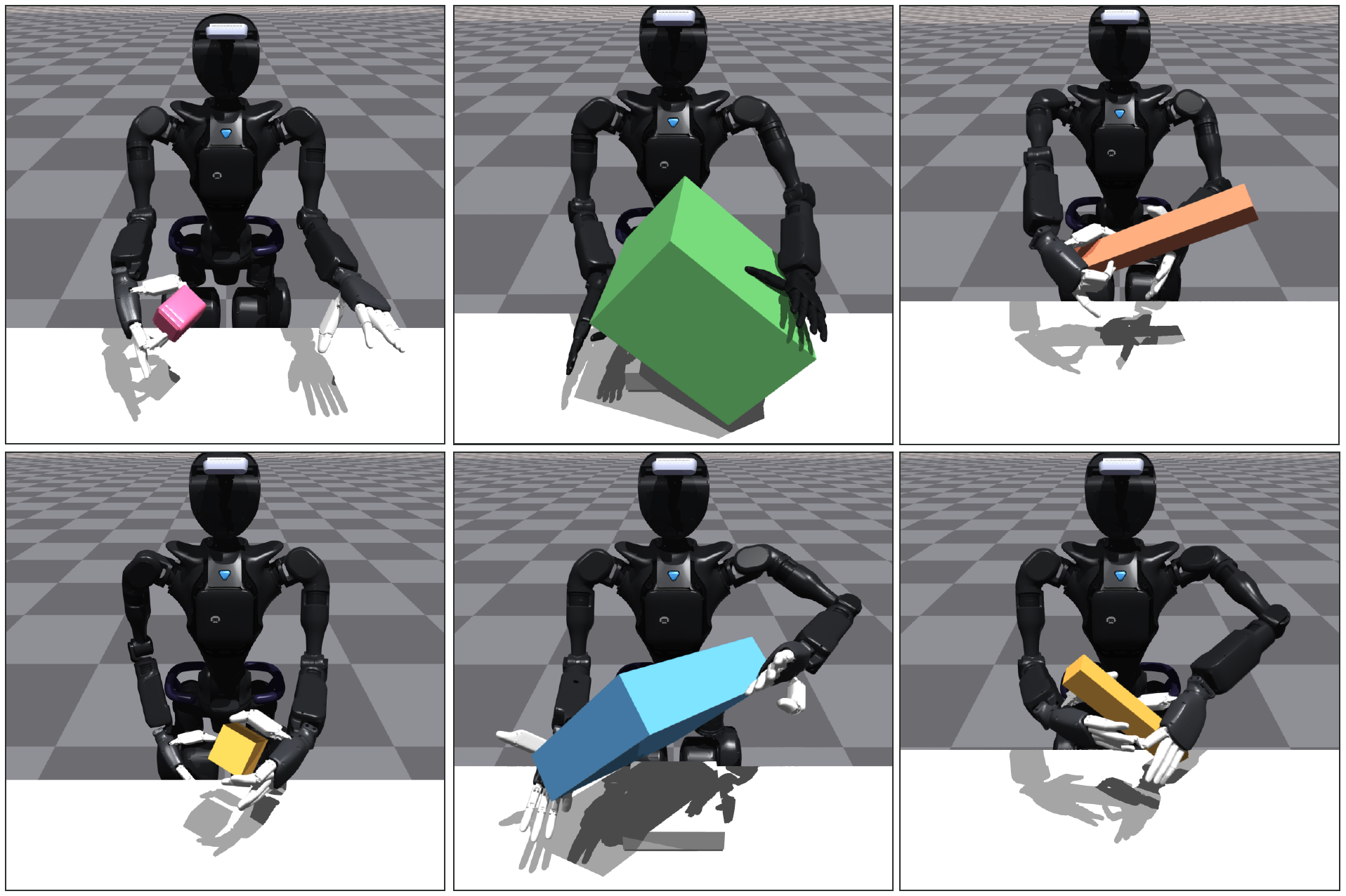}
    \caption{\small{\textbf{Policies learned in simulation.} Left: grasp-and-reach; middle: box lift; right: bimanual handover (right-to-left, left-to-right).}}
    \label{fig:sim}
  \end{minipage}
  \hfill
  \begin{minipage}[t]{0.62\textwidth}
    \includegraphics[width=\linewidth]{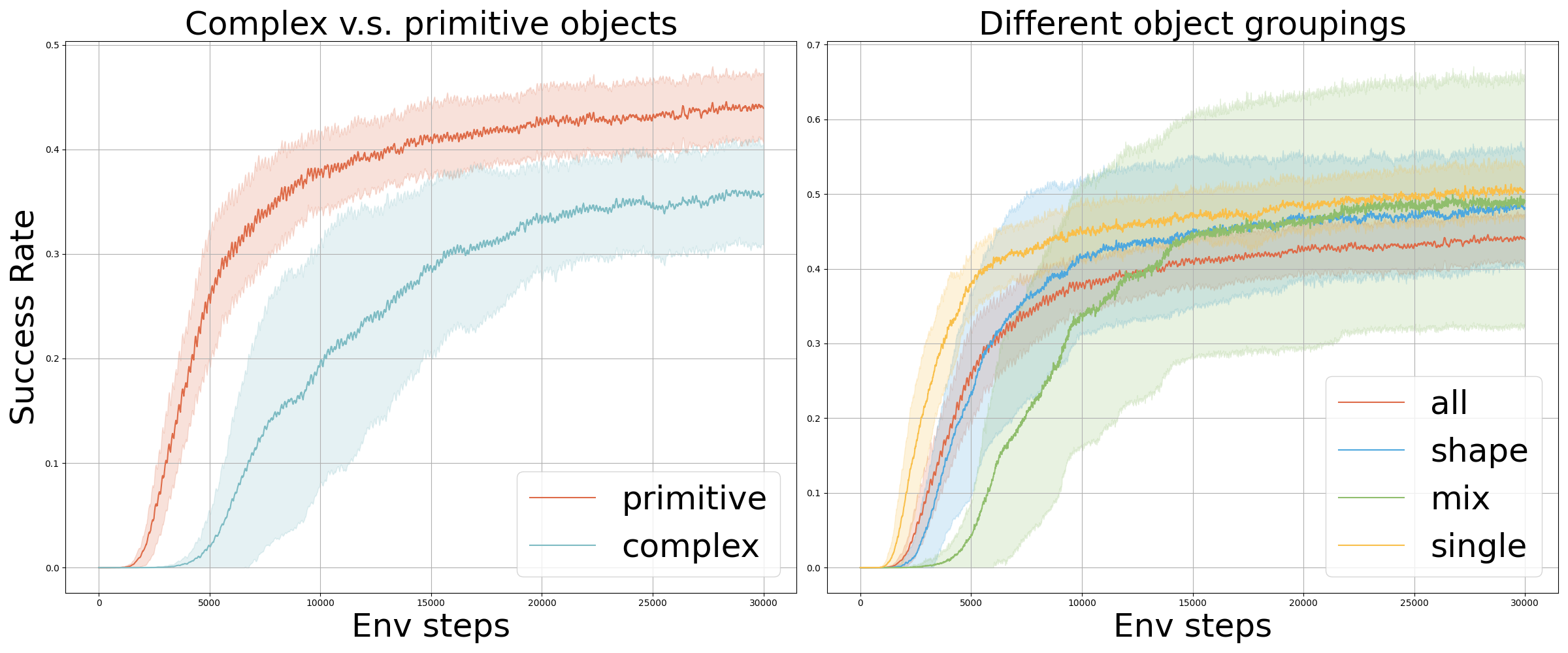}
    \caption{\small{\textbf{Training \texttt{grasp-and-reach} policy with different object sets.} Each curve is from 10 runs with different random seeds. Left: training with complex objects v.s. simple geometric primitive objects. Right: training with differently grouped geometric objects.
    }}
    \label{fig:objexp}
  \end{minipage}
\vspace{-1em}
\end{figure*}

\subsection{Task Definition}
\label{sec:task}
\vspace{-1em}

\textbf{(A) Grasp-and-reach.}
The robot must use one hand to grasp a tabletop object, lift it, and place it at a goal location. At initialization, a scripted vision module selects the closer hand to object. Test objects vary in shape, mass, volume, friction, color, and texture (see Figure~\ref{fig:teaser}). Each trial randomizes object pose and goal location.
\textbf{(B) Box lift.} The robot lifts a box too large for single-handed grasping. Box size, color, mass, and initial pose (with randomized position and yaw) are varied across trials.
\textbf{(C) Bimanual handover.} The robot grasps an object from one side of the table with one hand and hands it over to the other hand, which cannot reach the object directly. Objects vary in color, size, mass, and pose. We vary the initial pose of blocks in each trial.

\subsection{Evaluation of Real-to-Sim Modeling}
\vspace{-1em}

\begin{figure*}[t]
\centering
\begin{minipage}[b]{0.48\textwidth}
    \centering
    \includegraphics[width=\linewidth]{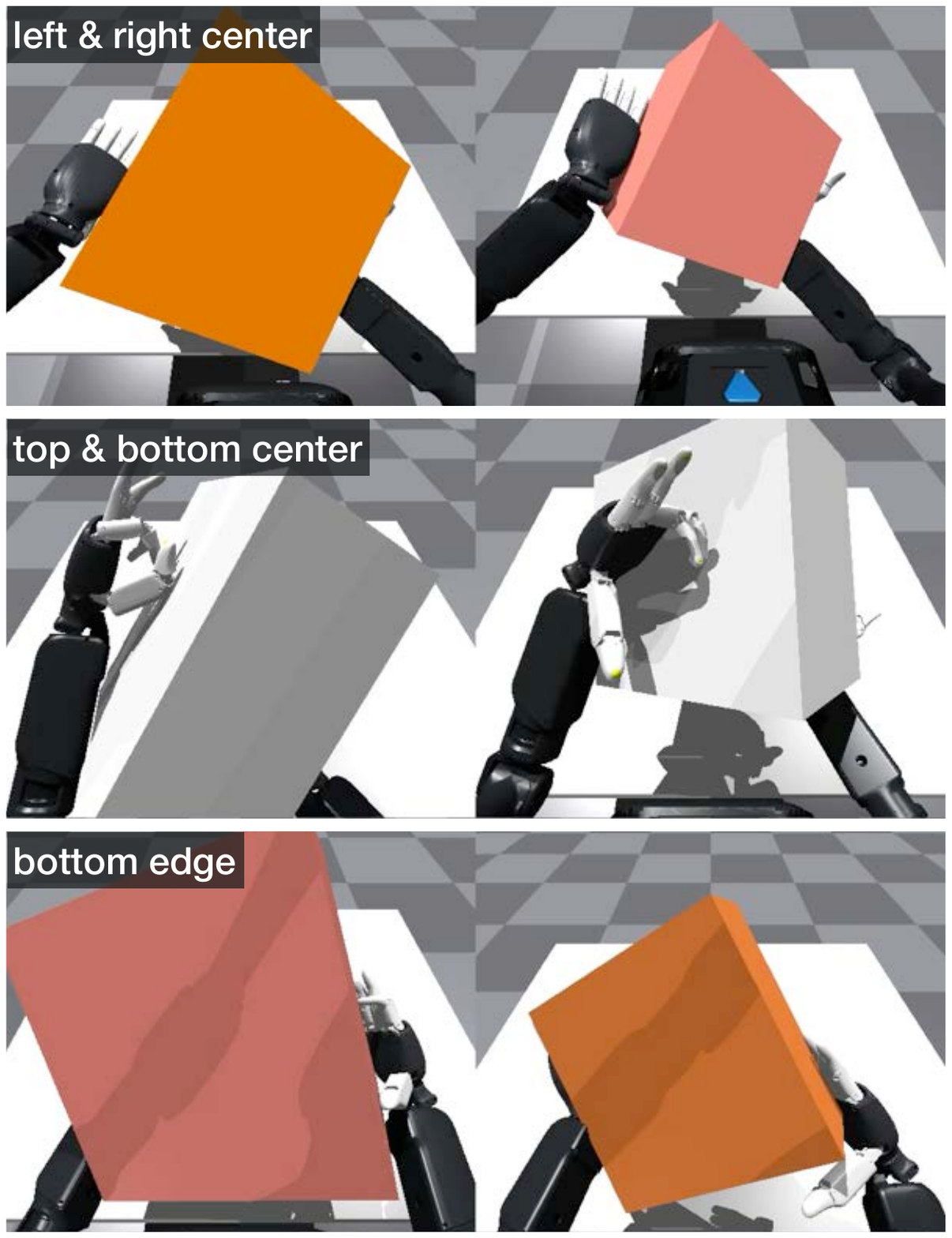}
    \caption{\small{\textbf{Different contact patterns emerge from different placements of contact markers.} Top: contact markers on the left and right side centers; middle: markers on the top and bottom side centers; bottom: markers on the bottom side edges.}}
    \label{fig:contactexp}
\end{minipage}
\hfill
\begin{minipage}[b]{0.5\textwidth}
    \centering
    \includegraphics[width=0.95\linewidth]{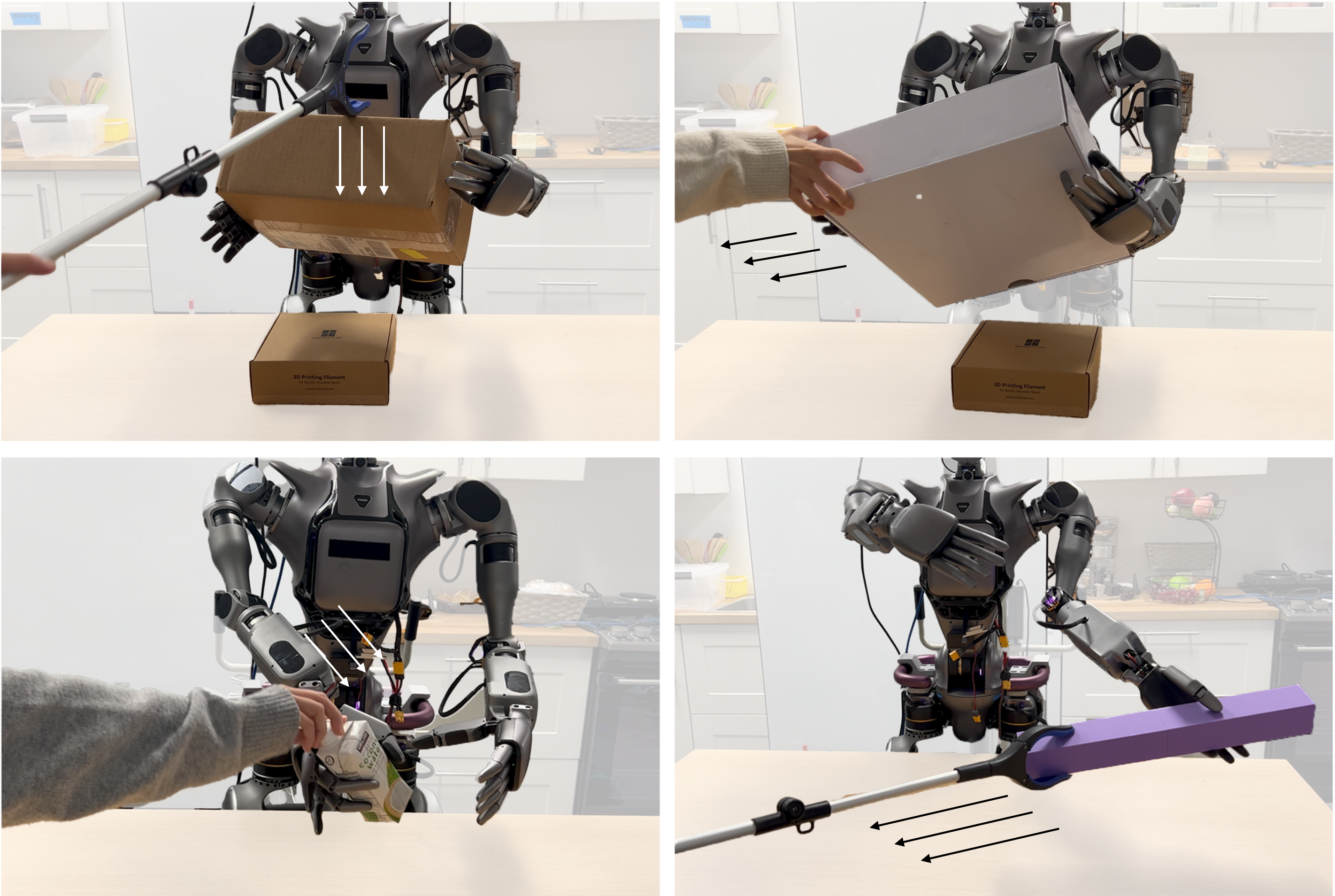}
    \caption{\small{\textbf{Policy robustness.} Our learned policies remain robust under different force perturbations, including knock (top left), pull (top right), push (bottom left), and drag (bottom right).}}
    \label{fig:robust}
    \vspace{0.5em} 

    {\setlength{\tabcolsep}{5pt}  
    \renewcommand{\arraystretch}{1.1}  
    \scriptsize  
    
    \begin{tabular}{lccc}
    \toprule
    \textbf{Autotune MSE} & \textbf{Lowest} & \textbf{Median} & \textbf{Highest} \\
    \midrule
    Grasp Success & 8 / 10 & 3 / 10 & 0 / 10 \\
    Reach Success & 7 / 10 & 3 / 10 & 0 / 10 \\
    \bottomrule
    \end{tabular}
    }
    \vspace{-0.3em}
    \captionof{table}{\small{\textbf{Lower MSE from autotune correlates with higher sim-to-real success rate.} For each set of modeling parameters, we test the sim-to-real transfer performance of 10 policy checkpoints (trained identically except for random seed). We evaluate success rate by stages on the \texttt{grasp-and-reach} task, and observe a correlation between lower MSE measured by autotune module and higher sim-to-real transfer success rate.}}
    \label{table:autotune}

\end{minipage}
\vspace{-1em}
\end{figure*}

\paragraphc{Effectiveness of autotuned robot modeling.}
We apply the autotune module described in Section~\ref{sec:realsim} to optimize the robot modeling parameters. To assess its effectiveness, we compare the sim-to-real transfer success rates of three sets of policy checkpoints, each trained with identical settings except for the robot modeling parameters. These parameter sets correspond to varying levels of modeling accuracy, as measured by the mean squared error (MSE) from autotune -- ranging from the lowest (i.e., smallest real-to-sim gap) to the highest (i.e., largest real-to-sim gap).  As shown in Table~\ref{table:autotune}, policies trained with autotuned models exhibit significantly better sim-to-real performance. Qualitative examples in our video further demonstrate successful transfer of \texttt{grasp-and-reach} policies to the Inspire hands, highlighting the generalizability of our autotune module.

\paragraphc{Effectiveness of approximate object modeling.} Empirically, we find that modeling objects as primitive geometric shapes (cylinders, cubes, and spheres) strikes a good balance between training efficiency and sim-to-real transferability. As shown in Figure~\ref{fig:objexp} (left), training \texttt{grasp-and-reach} policies with primitive shapes leads to faster convergence compared to using complex object geometries. More importantly, policies trained with randomized primitive shapes demonstrate strong generalization to a diverse set of unseen objects, as illustrated in our video.


\subsection{Evaluation of Reward Design}
\vspace{-1em}

\paragraphc{Task capabilities.} Enabled by our proposed reward design principle, a broad range of long-horizon, contact-rich tasks can be successfully solved using pure RL, as shown in Figure~\ref{fig:sim} and video. The resulting policies exhibit notable dexterity and robustness under various random force disturbances.

\paragraphc{Effectiveness of contact-based rewards.} In Figure~\ref{fig:contactexp}, we visualize how different contact behaviors emerge from varying the placement of contact markers, using the \texttt{box lift} task as an example. The contact stickers are procedurally generated along box sides or edges based on the box dimensions. The resulting behaviors closely reflect the specified contact positions, demonstrating the effectiveness of using contact markers to define contact goals.

\subsection{Evaluation of Policy Learning}
\vspace{-1em}

\paragraphc{Effectiveness of task-aware hand pose initialization.} In Table~\ref{table:humaninit}, we compare the percentage of successfully trained policies for each task with and without task-aware hand pose initialization. The results indicate that incorporating human priors at initialization significantly enhances exploration efficiency in challenging RL tasks.

\paragraphc{Divide-and-conquer distillation.} We evaluate our divide-and-conquer distillation strategy through two ablation studies. First, we examine how the granularity of task decomposition affects training efficiency in a multi-object \texttt{grasp-and-reach} task involving 10 objects. We compare four designs: (1) training a single policy on all objects (\textit{all}); (2) training three policies on shape-similar object groups (\textit{shape}); (3) training three policies on shape-diverse object groups (\textit{mix}); and (4) training ten separate single-object policies (\textit{single}). As shown in Figure~\ref{fig:objexp}, \textit{single} achieves the highest sample efficiency, followed by \textit{shape}, \textit{all}, and \textit{mix}. The average success rates also vary across designs, reflecting differences in task difficulty. Notably, policies trained on reduced object sets converge to similar final performance, while the \textit{all} policy consistently underperforms. Second, we evaluate the sim-to-real transfer success rate of each policy type on an in-distribution object over 30 trials. The \textit{mix} policy performs best (90.0\%), followed by \textit{shape} (63.3\%), \textit{single} (40.0\%), and \textit{all} (23.3\%). We hypothesize that the lower performance of \textit{single} and \textit{mix} arises from overfitting to specific geometries, while the poor performance of \textit{all} is consistent with its weaker RL training outcomes. These results suggest that divide-and-conquer distillation strikes a favorable balance between training efficiency and sim-to-real generalization.

\begin{table*}[t]
\centering

\begin{minipage}[b]{0.4\textwidth}
\centering
\setlength{\tabcolsep}{5pt}
\renewcommand{\arraystretch}{1.1}
\scriptsize
\begin{tabular}{lccc}
\toprule
\textbf{\% Success} & Grasping & Lifting & Handover \\
\midrule
with Human Init & 80\% & 90\% & 30\% \\
w/o Human Init & 60\% & 90\% & 0\% \\
\bottomrule
\end{tabular}
\caption{\small{\textbf{Initializing with human data.} Correlation between the percentage of successful task policies and whether human play data is used for initialization. We define \textit{successful} policies as those that achieve over 60\% episodic success during evaluation. For each task and each initialization setting, we test with 10 random seeds.}}
\label{table:humaninit}
\end{minipage}
\hfill
\begin{minipage}[b]{0.58\textwidth}
\centering
\setlength{\tabcolsep}{5.5pt}
\renewcommand{\arraystretch}{1.1}
\scriptsize
\begin{tabular}{lcccc}
\toprule
\textbf{Task} & Grasping & Lifting & HandoverA & HandoverB \\
\midrule
\multicolumn{5}{l}{\textbf{Depth + Pos}} \\
Pickup & 10 / 10 & 10 / 10 & 10 / 10 & 10 / 10 \\
Task Success & 10 / 10 & 10 / 10 & 9 / 10 & 5 / 10 \\
\midrule
\multicolumn{5}{l}{\textbf{Depth Only}} \\
Pickup & 2 / 10 & 0 / 10 & 0 / 10 & 0 / 10 \\
Task Success & 2 / 10 & 0 / 10 & 0 / 10 & 0 / 10 \\
\bottomrule
\end{tabular}
\caption{\small{\textbf{Comparing sim-to-real transfer performance between depth-and-position policy and depth-only policy.} We separate the \texttt{bimanual handover} task into two columns due to its longer horizon. Pickup success measures how often hands pick up the object. Combining 3D position with depth enables easier sim-to-real transfer.}}
\label{table:objrep}
\end{minipage}

\end{table*}

\subsection{Evaluation of Vision-Based Sim-to-Real Transfer}
\vspace{-1em}

\paragraphc{Effectiveness of mixing object representations.} We study the impact of different object representations on sim-to-real transfer performance, with results summarized in Table~\ref{table:objrep}. Our findings show that combining a dense representation (segmented depth image) with a sparse representation (3D object center-of-mass position) leads to improved transfer success. Notably, the performance gap between the combined depth-and-position policy and the depth-only policy widens for tasks where accurate understanding of full object geometry is more critical to success.

\subsection{System Capabilities}
\vspace{-1em}

\paragraphc{Task performance, generalization, robustness.} We evaluate the overall effectiveness of our system by reporting task success rates using the best-performing policy for each task. For each task, we perform 10 trials for each test object and compute the average success rate across all objects. We report a 62.3\% success rate for the \texttt{grasp-and-reach} task, 80\% for \texttt{box lift}, and 52.5\% for \texttt{bimanual handover}. To assess generalization, we test the \texttt{grasp-and-reach} policy on out-of-distribution objects and present qualitative evidence of successful zero-shot transfer in our video. Additionally, we evaluate the robustness of our policies under external force perturbations across all tasks, as shown in Figure~\ref{fig:robust} and our videos. More details on the object set for each task are reported in Figure~\ref{fig:teaser}.

\paragraphc{Extension to a more capable system.}
The learned RL policies can be seamlessly integrated with higher-level control structures such as finite state machines or teleoperation frameworks to enable longer-horizon task execution, while preserving dexterity, robustness, and generalization. As a proof of concept, our video showcases a general pick-and-drop system constructed by scripting sequences around the \texttt{grasp-and-reach} policy.

\section{Conclusion} 
\label{sec:conclusion}
\vspace{-1em}

We present a comprehensive recipe for applying sim-to-real RL to vision-based dexterous manipulation on humanoids. By addressing key challenges in environment modeling, reward design, policy learning, and sim-to-real transfer, we show that RL can be a powerful tool for learning highly useful manipulation skills without the need for extensive human demonstrations. Our learned policies exhibit strong generalization to unseen objects, robustness against force disturbances, and the ability to perform long-horizon contact-rich tasks.


\clearpage
\section{Limitations} 
\label{sec:limitations}

In this work, we investigate the key challenges in applying RL to robot manipulation and introduce practical and principled techniques to overcome the hurdles. Based on the techniques proposed, we build a sim-to-real RL pipeline that demonstrates a feasible path to solve robot manipulation, with evidence on generalizability, robustness, and dexterity.

However, the capabilities achieved in this work are still far from the kind of ``general-purpose'' manipulation that humans are capable of. Much work remains to be done to improve each individual component of this pipeline and unlock the full potential of sim-to-real RL.
For example, the reward design could be improved by integrating even stronger human priors, such as task demonstrations collected from teleoperation; alternative controller, such as torque controller, could also be explored.

There are also important open problems that our work does not address. For example, our work uses no novel technique to reduce the sim-to-real gap in dynamics other than applying naive domain randomization. We hypothesize that this could be a reason for the low success rate on \texttt{bimanual handover} task, which is the most dynamic among our collection of tasks.

Lastly, we find ourselves heavily constrained by the lack of reliable hardware for dexterous manipulation. While we use multi-fingered robot hands, the dexterity of these hands is far from that of human hands in terms of the active degrees of freedom.
We believe the dexterity of our learned policies is not limited by the approach, and we hope to extend our framework to robot hands with more sophisticated designs in the future.


\clearpage
\acknowledgments{We thank members of NVIDIA GEAR lab for help with hardware infrastructure, in particular Zhenjia Xu, Yizhou Zhao, and Zu Wang. This work was partially conducted during TL's internship at NVIDIA. TL is supported by NVIDIA and the National Science Foundation fellowship.}

\bibliography{references}  


\clearpage
\section*{Appendix}
\subsection{Environment Modeling Details}

\paragraphc{Modeling underactuated joints.}
Since modeling underactuated joint structure is not directly supported, we approximate the relationship between each pair of actuated and underactuated joints by fitting a linear function $q_{u} = k \cdot q_{a} + b$, where $q_u$ is the underactuated joint angle and $q_a$ is the actuated joint angle. Note that parameters $k, b$ are included as tunable parameters to search over using our autotune module detailed in Section~\ref{sec:realsim}.

\subsection{Reward Design Details}

We design generalizable rewards based on the principle outlined in Section~\ref{sec:reward} and list task reward details below.

Both \textbf{grasp} and \textbf{lift} tasks can be defined with the following goal states: (1) finger contact with the object; (2) the object being lifted up to a goal position. Our reward design can, therefore, follow by combining the contact goal reward and the object goal reward terms:
\begin{equation}
r(s_h,s_o) = r_{contact}(s_h, s_o) + r_{goal}(s_o)
\end{equation}
where $s_h$ includes fingertip positions, $s_o$ includes object center-of-mass position, and all contact marker positions (if any).

Similarly, the \textbf{handover} task can be defined with the following goal states: (1) one hand's finger contact with the object; (2) object being transferred to an intermediate goal position while still in contact with the first hand; (3) the second hand's finger contact with the object; (4) object being transferred to the final goal position. Due to the hand switching, we introduce a stage variable $a \in \{0,1\}$ and design the reward as follows:
\begin{equation}
\begin{split}
r(s_h,s_o) & = (1 - a) \cdot ( r_{contact}(s_{h_A}, s_{o_A}) + r_{goal}(s_{o_A})) \\
& + a \cdot (r_{contact}(s_{h_B}, s_{o_B}) + r_{goal}(s_{o_B}))
\end{split}
\end{equation}
where $s_{h_A}, s_{h_B}$ denote fingertip positions of the engaged hand at each stage, $s_o$ denote object center-of-mass position and desirable contact marker positions (if any) at each stage. At completion of each stage, we also reward policy with a bonus whose scale increases as stage progresses.

\subsection{Policy Training Details}

\paragraphc{RL implementation.}
To learn the specialist policies, the observation space includes object position and robot joint position at each time step, and the action space is robot joint angles. We use Proximal Policy Optimization~\cite{schulman2017proximal} with asymmetric actor-critic as the RL algorithm. In addition to the policy inputs, we provide the following privilege state inputs to the asymmetric critic: arm joint velocities, hand joint velocities, all fingertip positions, object orientation, object velocity, object angular velocity, object mass randomization scale, object friction randomization scale, and object shape randomization scale. Both the actor and critic networks are 3-layer MLPs with units $(512,512,512)$.

\paragraphc{Domain randomization.}
\label{sec:dr}
Physical randomization includes the randomization of object friction, mass, and scale. We also apply random forces to the object to simulate the physical effects that are not implemented by the simulator. Non-physical randomization models the noise in observation~(e.g. joint position measurement and detected object positions) and action. A summary of our randomization attributes and parameters is shown in Table~\ref{table:dr}.
 
\begin{table}[!t]
\renewcommand\arraystretch{1.05}
\caption{Domain Randomization Setup.}
\centering
\begin{tabular*}{\linewidth}{l@{\extracolsep{\fill}}c}
\toprule
Object: Mass~(kg)             & [0.03, 0.1]    \\
Object: Friction              & [0.5, 1.5]     \\
Object: Shape                 & $\times\mathcal{U}(0.95, 1.05)$     \\
Object: Initial Position~(cm) &  $+\mathcal{U}(-0.02, 0.02)$ \\
Object: Initial $z$-orientation & $+\mathcal{U}(-0.75, 0.75)$ \\
Hand: Friction                & [0.5, 1.5]    \\
\midrule
PD Controller: P Gain         &  $\times\mathcal{U}(0.8, 1.1)$      \\
PD Controller: D Gain         &  $\times\mathcal{U}(0.7, 1.2)$     \\
\midrule
Random Force: Scale           & 2.0       \\
Random Force: Probability     & 0.2    \\
Random Force: Decay Coeff. and Interval & 0.99 every 0.1s     \\ 
\midrule
Object Pos Observation: Noise & 0.02      \\
Joint Observation Noise.      & $+\mathcal{N}(0, 0.4)$     \\
Action Noise.                 & $+\mathcal{N}(0, 0.1)$   \\
\midrule
Frame Lag Probability         & 0.1 \\
Action Lag Probability        & 0.1 \\
\midrule
Depth: Camera Pos Noise~(cm)      & 0.005  \\
Depth: Camera Rot Noise~(deg)      & 5.0  \\
Depth: Camera Field-of-View~(deg)  & 5.0  \\
\bottomrule
\end{tabular*}
\label{table:dr}
\end{table}

\subsection{Distillation Details}

To learn the generalist policy, we reduce the choices of observation inputs to the robot joint states and selective object states, including 3D object position and egocentric depth view, since privileged information is unavailable for sim-to-real transfer. To more efficiently utilize the trajectory data and improve training stability, for each sub-task specialist policy, we evaluate for 5000 steps over 100 environments, saving trajectories filtered by success at episode reset on the hard disk. We then treat the saved data as ``demonstrations'' and learn a generalist policy for each task with Diffusion Policies~\cite{chi2023diffusion}.

The proprioception and object position states are concatenated and passed through a three-layer network with ELU activation, hidden sizes of $(512,512,512)$, and an output feature size of $64$. For depth observations, we use the ResNet-18 architecture~\cite{he2016deep} and replace all the BatchNorm~\cite{ioffe2015batch} in the network with GroupNorm~\cite{wu2018group}, following~\cite{chi2023diffusion}. All the encoded features are then concatenated as the input to a diffusion model. We use the same noise schedule (square cosine schedule) and the same number of diffusion steps (100) for training as in \cite{chi2023diffusion}.
The diffusion output from the model is the normalized 7 DoF absolute desired joint positions of each humanoid arm and the 6 DoF normalized ($0$ to $1$) desired joint positions of each humanoid hand. We use the AdamW optimizer~\cite{kingma2014adam, loshchilov2017decoupled} with a learning rate of $0.0001$, weight decay of $0.00001$, and a batch size of 128. Following \cite{chi2023diffusion}, we maintain an exponential weighted average of the model weights and use it during evaluation/deployment.

\end{document}